\documentclass[letterpaper, 10 pt, conference]{ieeeconf}  

\IEEEoverridecommandlockouts                              

\usepackage{amsmath, amssymb, mathtools}
\usepackage{xcolor}
\usepackage{graphicx}
\usepackage{subfigure}
\usepackage{caption}
\usepackage{svg}
\makeatletter
\let\NAT@parse\undefined
\makeatother
\usepackage[colorlinks=true, citecolor=green, linkcolor=red]{hyperref}  
\usepackage{cite}
\usepackage{algorithm}
\usepackage{algorithmic}

\usepackage{multirow}
\usepackage{booktabs}
\usepackage[T1]{fontenc}

\title{\LARGE \bf
VOLoc: Visual Place Recognition by Querying Compressed Lidar Map
}

\author{Xudong Cai, Yongcai Wang, Zhe Huang, Yu Shao and Deying Li 
\thanks{All authors are with the Department of Computer Science, School of Information, Renmin University of China, Beijing 100872, China. Corresponding author: Yongcai Wang. \{xudongcai, ycw, huangzhe21, 2017202116, deyingli\}@ruc.edu.cn}
\thanks{Dr. Li is supported in part by the National Natural Science Foundation of China Grant No. 12071478. Dr. Wang is supported in part by the National Natural Science Foundation of China Grant No. 61972404, Public Computing Cloud, Renmin University of China, and the Blockchain Lab. School of Information, Renmin University of China.}
}

\begin{document}

\maketitle
\thispagestyle{empty}
\pagestyle{empty}

\begin{abstract}
The availability of city-scale Lidar maps enables the potential of city-scale place recognition using mobile cameras. However, the city-scale Lidar maps generally need to be compressed for storage efficiency, which increases the difficulty of direct visual place recognition in compressed Lidar maps.
This paper proposes VOLoc, an accurate and efficient visual place recognition method that exploits geometric similarity to directly query the compressed Lidar map via the real-time captured image sequence. 
In the offline phase, VOLoc compresses the Lidar maps using a \emph{Geometry-Preserving Compressor} (GPC), in which the compression is reversible, a crucial requirement for the downstream 6DoF pose estimation. 
In the online phase, VOLoc proposes an online Geometric Recovery Module (GRM), which is composed of online Visual Odometry (VO) and a point cloud optimization module, such that the local scene structure around the camera is online recovered to build the \emph{Querying Point Cloud} (QPC). 
Then the QPC is compressed by the same GPC, and is aggregated into a global descriptor by an attention-based aggregation module, to query the compressed Lidar map in the vector space.
A transfer learning mechanism is also proposed to improve the accuracy and the generality of the aggregation network. 
Extensive evaluations show that VOLoc provides localization accuracy even better than the Lidar-to-Lidar place recognition, setting up a new record for utilizing the compressed Lidar map by low-end mobile cameras. The code are publicly available at \href{https://github.com/Master-cai/VOLoc}{https://github.com/Master-cai/VOLoc}. 
\end{abstract}

\vspace{-3 ex}
\section{INTRODUCTION}
Visual place recognition (VPR) identifies the most likely map segment that the camera is positioning within, which generally serves as the initialization step, i.e., \emph{global localization},  for achieving ultimate accurate 6DoF pose estimation based on the camera captured images~\cite{HLoc}.
It is a crucial problem in autonomous driving, augmented reality, and robot systems. 
Traditional VPR primarily relies on Image-to-Image query, which uses image databases attached with Geo-information as the map~\cite{torii201524}. Then the place recognition is accomplished by querying the image database using the real-time captured images~\cite{DBLP:conf/cvpr/BrachmannR18, arandjelovic2016netvlad, DBLP:journals/ral/XieDWLWTZ22, sattler2016efficient}.
However, the image database itself suffers low accuracy and poor robustness issues for appearance changes (e.g., view angle, illumination, season)~\cite{toft2020long}.

\begin{figure}[t]
  \centering
  \includegraphics[width=0.9\linewidth]{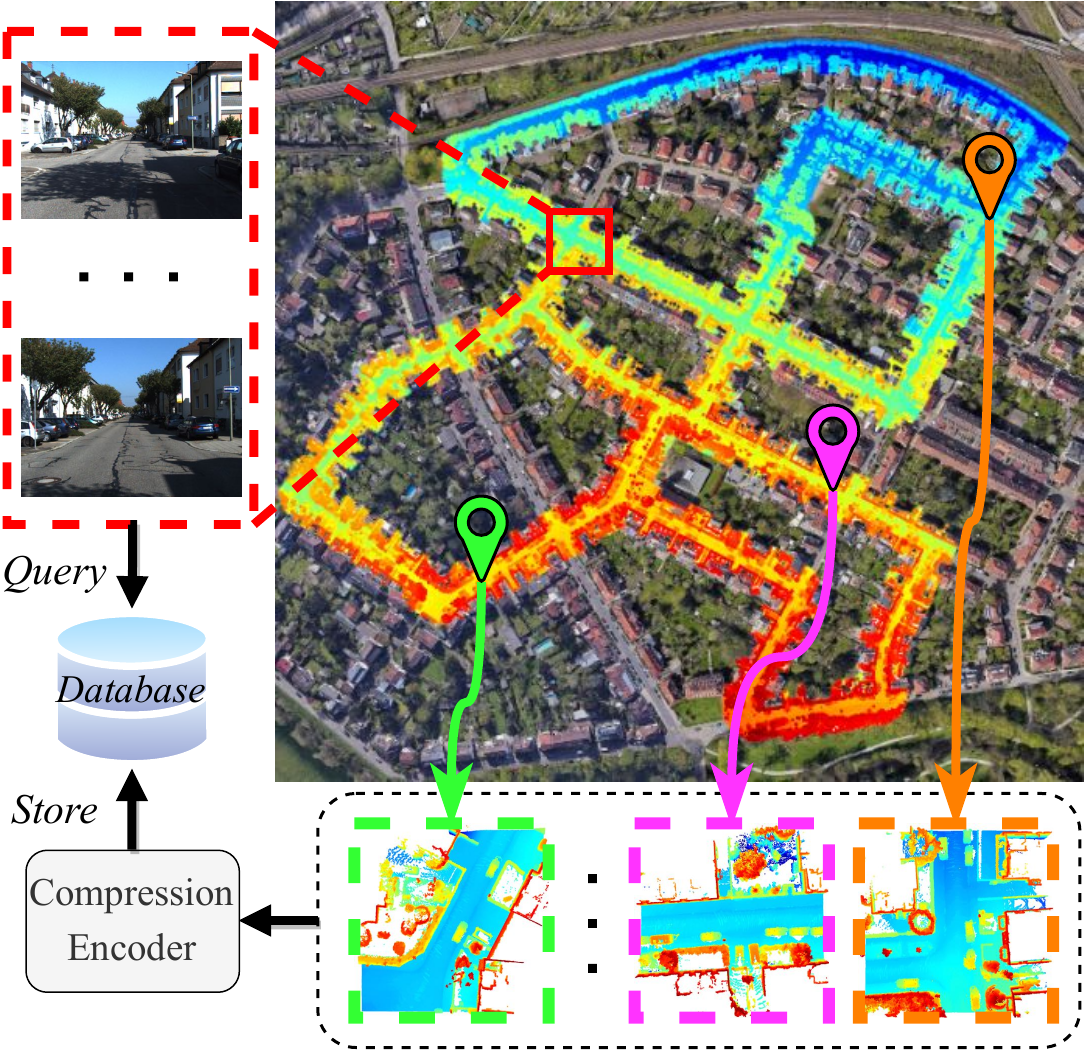}
  \caption{Location images in compressed Lidar maps}
  \label{fig:intro}
\end{figure}

Instead, the rapid growth of Lidar-based road surveying for constructing high-resolution 3D maps to support autonomous driving~(e.g., Nuscenes~\cite{Nuscenes}, Waymo~\cite{Waymo}) has led to the increasing accessibility of city-scale Lidar maps. The Lidar maps are more robust against the weather and illumination changes. This inspires the interest of Image-to-Lidar place recognition~\cite{cattaneo2020global}, that localizes where the images are taken in the Lidar maps. However, as the city-scale point cloud needs huge storage consumption, compression is generally adopted~\cite{DBLP:conf/cvpr/YangSLL0CT22} to efficiently store the city-scale Lidar maps. The compression exacerbates the modal gap and the difficulty of Image-to-Lidar place recognition. 

In this work, our goal is to conduct VPR directly in the compressed Lidar maps, as illustrated in Figure~\ref{fig:intro}. It is challenging because of two reasons.
At first, the images and the point clouds have different nature and directly matching between them is difficult. Current Image-to-Lidar place recognition methods project images and point clouds into intermediate representations, such as Bird's Eye View (BEV) images \cite{I2P-Rec},  or a shared vector space \cite{cattaneo2020global}.
However, the former only uses a single frame image, which suffers from limited field-of-view and loses altitude information. The latter neglects geometric information and reports unsatisfactory performances, whose Recall@1 is only around 40\% for Image-to-Lidar localization~\cite{cattaneo2020global}. Secondly, the compression of Lidar maps further increases the modal gap. Existing works have proposed some Lidar-to-Lidar place recognition methods, mainly using uncompressed maps~\cite{DBLP:conf/mir/SunL0F020, uy2018pointnetvlad}. Only a few researchers exploit to query the compressed Lidar map using Lidar-based query\cite{DBLP:conf/cvpr/WeiBWMU19, wiesmann2022retriever}. But how to use images to directly query compressed Lidar maps remains unexplored.

In this paper, we propose VOLoc, a novel framework that exploits \emph{geometrical similarity} to solve the challenges of Image-to-Lidar place recognition without decompressing the Lidar maps. 
The key idea is to utilize geometric information as an intermediate representation to close the modality gap.
On one hand, we exploit a Geometry-Preserving Compressor (GPC) to compress the segmented Lidar maps, which serve as the location \emph{database}. Notably, GPC compresses the point clouds by clustering and downsampling, which preserves the geometric structure and ensures the compression is reversible. The reversible compression is critical for the downstream precise  6DOF pose estimation.  Then an \emph{attention-based aggregation module} is proposed to convert the compressed sub-maps into global descriptors to integrate the neighboring information for the ease of querying. 

In online phase, the local geometric structure around the camera is rebuilt through the online Geometric Recovery Module (GRM), which comprises a Visual Odometry (VO) module~\cite{engel2017direct, campos2021orb, qin2018vins} and a point cloud optimization module. GRM strives to recover as much local structure information as possible, and outputs the reconstructed point cloud as Querying Point Cloud (QPC). Then the QPC is compressed by the same GPC and aggregated into a querying global descriptor by the same aggregation module. Then the location index with the closest vector distance in the database is returned as the place recognition result. 

We pre-train the aggregation network on a large Lidar point cloud dataset and fine-tune it on the VO-generated point clouds. 
The pre-training gives the network prior geometric knowledge, enhancing robustness and generalization. A transformer variant~\cite{jaegle2021perceiver} is adopted in the querying network to reduce the computational cost.
To validate VOLoc, we explore three VO systems, i.e., DSO~\cite{engel2017direct}, ORB-SLAM3~\cite{campos2021orb}, and VINS-Mono~\cite{qin2018vins}.
Experiments on the KITTI dataset~\cite{Geiger2013VisionMR} demonstrate our method offers comparable accuracy to that of the state-of-the-art Lidar-to-Lidar place recognition methods.
The key contributions are: 

(1) Geometric similarity is explored to enable Image-to-Compressed Lidar place recognition. 

(2) Geometry-Preserving Compressor (GPC) is exploited  to build the database and a Geometric Recovery Module (GRM) is proposed to recover the local geometric information from the image sequence. 

(3) A transfer learning scheme is proposed to train the aggregation module, that greatly boosts the accuracy. 

(4) A Visual-to-Lidar localization dataset based on KITTI is constructed for evaluating the proposed method and for the use in society. 

\section{Related Work}

\subsection{Image-to-Image place recognition}

Image-to-image place recognition uses images to query the image database, which can be further divided into two categories: \emph{local feature based methods} and \emph{global appearance based methods}. 
Local feature based methods firstly extracts local features (SIFT~\cite{lowe2004distinctive}, SuperPoint~\cite{detone2018superpoint}) and then aggregated them into global descriptors via Bag-of-Words (BoW)~\cite{sivic2003video} or Vector of Locally Aggregated Descriptors (VLAD)~\cite{jegou2010aggregating}.  The global appearance based methods~\cite{berton2022rethinking, APANet, DMPCANet} usually use a classic network (VGG~\cite{vgg} or ResNet~\cite{resnet}) as the backbone to extract global appearance descriptors~\cite{arandjelovic2016netvlad}, and then are trained by contrastive learning using GNSS information as a weak supervision signal~\cite{hausler2021patch, jin2017learned, peng2021semantic}.
In both categories, the image-based query is carried out by calculating the similarity between the global descriptors.

\subsection{Lidar-to-Lidar place recognition}
Lidar-to-Lidar methods use local Lidar point clouds to query large Lidar maps. Scan Context series~\cite{ScanContext,ScanContext++,DiSCO} project point clouds to BEV to identify locations. BEVPlace~\cite{BEVPlace} uses group convolution to extract rotation equivariant local features from bird's-eye-view~(BEV) images.
PointNetVLAD~\cite{uy2018pointnetvlad} is the first work that uses PointNet~\cite{qi2017pointnet} to extract local features from point cloud and aggregates them into global descriptors. Subsequent works improve descriptor generation and discriminability by enhancing local features~\cite{liu2019lpd};  assigning different weights to each point during feature aggregation~\cite{zhang2019pcan}; considering the geometric relationship among points~\cite{sun2020dagc}; or utilizing 3D convolutions on sparse voxelized point clouds~\cite{Minkloc3d, MinkLoc3Dv2}. OverlapTransformer~\cite{OverlapTransformer} applies the attention mechanism for better performance. 
Although these methods generally perform well, storing the origin point clouds needs large storage space. To alleviate this problem, Wiesmann ~\cite{wiesmann2022retriever} proposes a novel attention-based aggregation module to localize the Lidar sub-map directly in the compressed map without decompressing.

\subsection{Visual-to-Lidar place recognition}
\label{related work:cross-model methods}
Localizing an image in the point cloud maps is far unexplored. The different nature of images and point clouds makes the cross-modal query challenging. 
Some works project point clouds into images~\cite{wolcott2014visual, cattaneo2019cmrnet} or reconstruct point clouds~\cite{caselitz2016monocular,li20203d} from images to mitigate the modal gap, but all of them need a rough pose estimation. 2D3D-MatchNet~\cite{feng20192d3d} is the first work that tries to solve the place recognition problem in a metric learning way. They feed detected SIFT features~\cite{lowe2004distinctive} in images and Intrinsic Shape Signatures (ISS) features~\cite{zhong2009intrinsic}  in point clouds into a network to project them to the same embedding space to calculate similarity.
LCD~\cite{pham2020lcd} follows the same pipeline, but uses learned features instead of the traditional features. 
Cattaneo et al.~\cite{cattaneo2020global} directly put the whole images and point clouds into neural networks to create a shared embedding space. I2P-Rec~\cite{I2P-Rec} projects both the point clouds and the images into BEV images to close the modal gap.
But these methods provide unsatisfactory locating accuracies since the single image suffers from limited field-of-view and doesn't well correlate with the point cloud.
The problem of Image-to-Compressed-Lidar map place recognition remains unexplored.

\begin{figure*}[ht]
  \centering
  \includegraphics[width=0.8\linewidth]{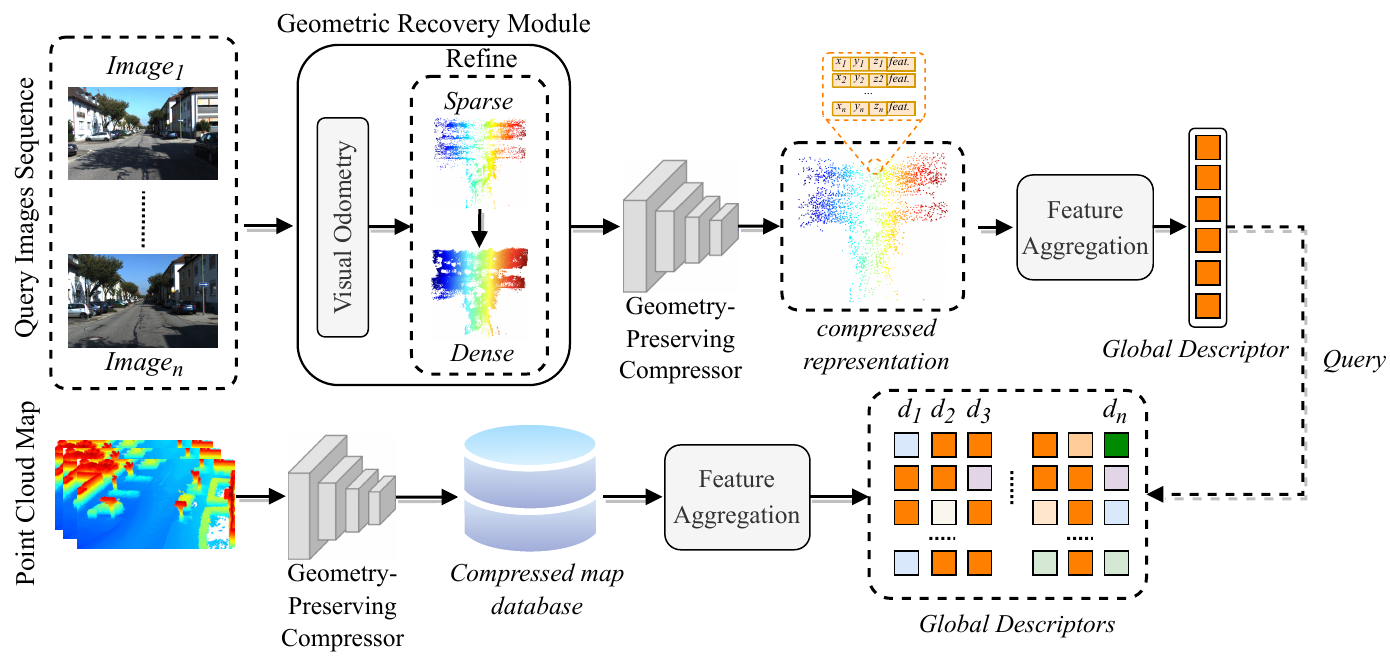}
  \caption{Overall framework of VOLoc}
  \label{fig:Overall}
\vspace{-2 em}
\end{figure*}


\section{Proposed Method}
\subsection{Problem Description}
Consider a city-scale point cloud map $\mathcal{M}$ which is collected along the city roads. The map is segmented into  segments of equal size. We compress the segmented maps for storage efficiency and setup a database,  i.e., $\mathcal{DB}=\{c_1, c_2, ..., c_N\}$, where $c_i$ is the $i$th compressed segment. A client equipped with a mono-camera queries the database using its captured images to find which segment the client is most possibly located at.

\subsection{Method Overview}
The overview of the proposed method is shown  in Figure~\ref{fig:Overall}. 
The Lidar sub-maps are first processed by Geometry-Preserving Compressor (Section~\ref{sec:GPC}),  and are then processed by the Feature Aggregation module (Section~\ref{sec:Global Feature Aggregation}) to be converted into global descriptors $D_d=\{d_1, d_2, ..., d_N\}$.

The query images go through the Geometric Recovery Module (Section~\ref{sec:GRM}) and the same GPC to generate the compressed query point cloud,  which is then converted into querying global descriptors $d_q$, using the same feature aggregation module (Section~\ref{sec:Global Feature Aggregation}). 
Place recognition is then done by retrieving the most similar descriptor in $D_d$ with $d_q$. 
A combined loss~(Section~\ref{sec:loss}) and a transfer learning scheme are applied to train the aggregation module (Section~\ref{sec:Transfer Learning}).

\subsection{Geometry-Preserving Compressor}
\label{sec:GPC}
To preserve geometric properties while reducing storage, the Geometry-Preserving Compressor (GPC) uses a grid-downsampling compressor~\cite{wiesmann2021deep} with an autoencoder design.

The encoder has three KPConv-downsampling blocks. Each block uses kernel point convolutions (KPConv)~\cite{KPConv} to aggregate features and uses grid-based downsampling to reduce the number of points. KPConv directly operates on the points and uses the points sampled from their neighbors as convolutional kernels to learn local geometric features. Grid-based downsampling is more efficient than the widely used Furthest Point Sampling (FPS) and keeps an even point distribution rather than losing most points in sparser areas. 
 The encoder compresses a $N$-points sub-map  $m_i \in \mathbb{R}^{N\times 3}$ into a compressed sub-map $c_i: \mathbb{R}^{N\times 3} \rightarrow \mathbb{R}^{N_c \times 6}, N_c \ll N$. The grid sizes of downsampling are $0.1m, 0.5m, 1m$, corresponding to the first to the third layer. Each 3D point in a compressed map is associated with a 3-dimensional feature. The decoder uses compressed maps to rebuild the origin point clouds by four deconvolutional blocks. We pre-train the autoencoder on the KITTI dataset in a self-supervised way as described in~\cite{wiesmann2021deep}. We only use the encoder part as the compressor and freeze its weights.

\subsection{Geometric Recovery Module}
\label{sec:GRM}
On the query end, we recover the geometric structure of the input images via the Geometric Recovery Module (GRM). In GRM, the images are first processed by Visual Odometry~(VO) to output sparse visual point clouds, which are further refined by filtering and densifying for better recovering geometry.

\paragraph{Visual Point Cloud Reconstruction}
Given a series of images, VO is exploited to track the trajectory and recover the sparse visual point clouds. 
Three most representative VO, i.e., Direct Sparse Odometry (DSO)~\cite{engel2017direct}, VINS-Mono~\cite{qin2018vins}, and ORB-SLAM3~\cite{campos2021orb} are employed to show our method is applicable to various VO methods. 
The rebuilt point clouds are divided by separating the estimated trajectories into a certain interval, aiming for a similar coverage area as the sub-maps in the database.

\paragraph{Visual Point Cloud Refine}
The rebuilt Visual Point Clouds are sparse, noisy and unevenly distributed. To optimize them, we propose a simple but efficient method to filter outliers and complete the Visual Point Clouds.

We first filter outliers caused by the VO systems as they may introduce undesired noise. In particular, the points that are far away from their neighbors compared to the average neighbor distance are removed. 
For each point $p_i$, its $K$ nearest neighbors (using KDTree) are found and the mean distance $t_i$ between these neighbors to $p_i$ is calculated. The global mean distance $T_m$ and the standard deviation $\sigma$ are derived from all $t_i$ values. A point $p_i$ is classified as an outlier when $t_i \geq T_m + \mu \times \sigma$. We set $K=20$ and $\mu=2$ in practice.
Then an interpolation method is used to densify the Visual Point Clouds. For each point $p_i$, the top $K$ closest neighbors $\{p^i_n | n \leq K\}$ are retrieved. We insert a new point $p = \frac{p_{i}+p_n^{i}}{2} $ between $p_i$ and every neighbor point $p^i_n$. To balance the time cost and the densifying effect, we set $K=10$ for point clouds built by DSO and $K=20$ for point clouds built by ORB-SLAM and VINS-Mono.
To eliminate the impression of scale uncertainty, we normalized the scale of point clouds to $[0, 1]$. The refined point clouds work as the Query Point Clouds (QPCs) and are compressed and aggregated into global descriptors by the following Global Feature Aggregation module.

\vspace{-1 ex}
\subsection{Global Feature Aggregation}
\label{sec:Global Feature Aggregation}

Directly querying compressed QPC in the database is still challenging. 
Traditional methods~\cite{uy2018pointnetvlad} aggregate all points into a global descriptor and calculate the similarity. 
However, it treats good and noisy points equally, introducing noise and degrading discriminability.  
To address this, we use an attention-based aggregation network. 
The core idea is to use the attention mechanism to focus on informative and accurate points that can better depict the geometric structure. The global receptive field of the attention mechanism can also enhance the expressiveness of global descriptors.

\begin{figure}[h]
\vspace{-1.3 ex}
  \centering
  \includegraphics[width=0.9\linewidth]{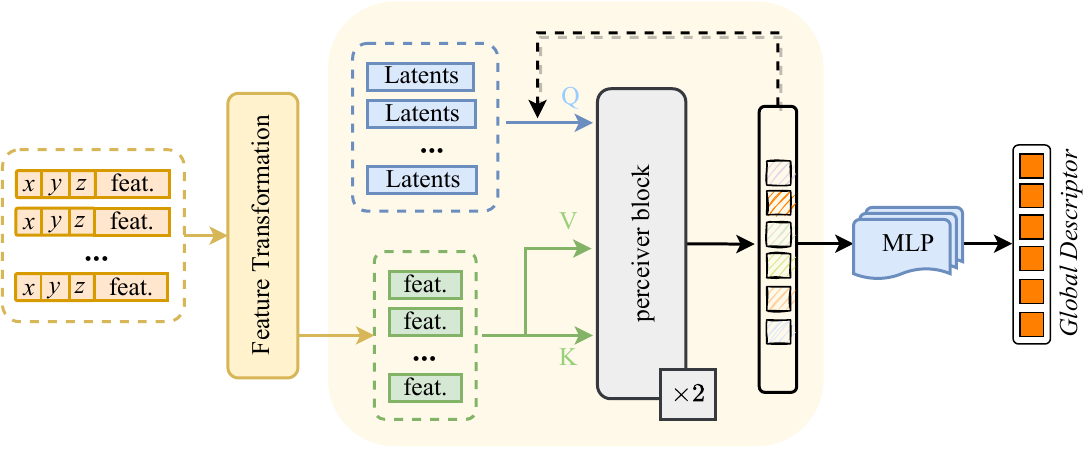}
  \caption{Architecture for Global Feature Aggregation}
  \label{fig:Network Architecture}
  \vspace{-1 em}
\end{figure}

Figure~\ref{fig:Network Architecture} shows the network's structure. We first utilize a T-Net variant~\cite{qi2017pointnet} to transform all input points to $F_a$, which are in a common feature space. This feature transformation makes $F_a$ invariant to the viewpoint changes and is more suitable for global descriptor aggregation.

The original attention mechanism projects entire feature sequences to $Q \in \mathbb{R}^{N_c \times D}$, $K \in \mathbb{R}^{N_c \times D}$, and $V \in \mathbb{R}^{N_c \times D}$. Yet, the quadratic growth of the attention score matrix $W \in \mathbb{R}^{N_c \times N_c}$ computed by the product of $Q$ and $K$ is computationally demanding. To address this, we employ perceiver-like attention~\cite{jaegle2021perceiver}. We substitute $Q$ with randomly initialized and fixed-size latent vectors $T \in \mathbb{R}^{N_t \times D}$, with $N_t \ll N_c$. The latent vectors $T$ will be optimized during training. 
Since $N_t$ is fixed and far less than $N_c$, the computational complexity decreases from $\mathcal{O}(N_c^2)$ to $\mathcal{O}(N_c)$.

The Perceiver block includes a cross-attention layer and four self-attention layers. It takes transformed features $F_a$ and latent vectors $T$ as input. The former is treated as $K$ and $V$ and the $T$ is treated as $Q$. Each block outputs the latent features, which will be treated as $Q$ to go through the next Perceiver block, resulting in the final latent features. The latent features are fed into an MLP to generate the global descriptors. We use two Perceiver blocks in the network. 

\vspace{-1 ex}
\subsection{Combined Loss Function}
\label{sec:loss}
This section describes the loss function used to train our network.
We denote $d_q$ as the anchor $P_a$; descriptors $d_i \in D_d$ describing the same place make up the positives set $S_{pos} = \{ P_1^{pos}, P_2^{pos}, ..., P_N^{pos} \}$; and those representing different places set up the negatives set $S_{neg}=\{ P_1^{neg}, P_2^{neg}, ..., P_N^{neg} \}$. Our target is to make the distance between $P_a$ and $P_i^{pos}$ 
be much less than the distance between $P_a$ and $P_i^{neg}$.
We use the Lazy quadruplet loss~\cite{uy2018pointnetvlad} in our network. The loss is defined as: 
\begin{align}
\begin{aligned}
		L_{VtoL} = max(d(P_a, P_h^{pos})-d(P_a, P_h^{neg})+\alpha , 0)\\ + max(d(P_a, P_h^{pos})-d(P_h^{neg}, P_s^{neg})+\beta, 0)
\end{aligned}
\end{align}
We take the Euclidean distance between descriptors as the distance function $d(\cdot)$. The $P_h^{pos}$ is the hardest positive sample in $S_{pos}$ and the $P_h^{neg}$ is the hardest negative sample in $S_{neg}$. The $P_s^{neg}$ means the second negative sample is far away from the anchor $P_a$ and other negatives. 

To make the descriptors more discriminative, we expect the distances among descriptors aggregated from the QPCs to obey the same rules. Likewise, we use the Lazy quadruplet loss to constrain them. The loss is defined as:
\begin{align}
\begin{aligned}
		L_{VtoV} = max(d(P_a, P_h^{vpos})-d(P_a, P_h^{vneg})+\alpha , 0)\\ + max(d(P_a, P_h^{vpos})-d(P_h^{vneg}, P_s^{vneg})+\beta, 0)
\end{aligned}
\end{align}
The definition of $P_h^{vpos}$, $P_h^{vneg}$, $P_s^{vneg}$ are similar with $P_h^{pos}$, $P_h^{neg}$, $P_s^{neg}$. The only difference is that the descriptors they stand for are aggregated from QPC instead of the database.

The final loss $L$ is defined as:
\begin{align}
L = L_{VtoL} + L_{VtoV}	
\end{align}

\subsection{Transfer Learning}
\label{sec:Transfer Learning}
To make the descriptors learned from the QPC more in line with those learned from the database, we propose a transfer learning approach to pre-train the network on a large Lidar point cloud dataset and then fine-tune it on the QPCs.

We first train our Feature Aggregation network on Oxford Robotcar dataset~\cite{maddern20171}, a large Lidar dataset including about 30000 point clouds. During the pre-training, we only use $L_{VtoL}$ in the loss function. This pre-training allows the model to learn general and robust geometric features. Subsequently, we fine-tune the pre-trained network on the QPCs. To enable the network to learn task-specific features while leveraging the general features learned from the Lidar data, we set the learning rate of the last MLP layers 10 times larger than the rest of the network. By using transfer learning, we leverage the knowledge learned from Lidar point clouds, improving the robustness of our network in handling the differences between Lidar and VO-generated point clouds.


\section{Experimental Evaluation}
\begin{table*}[t]
 \caption{Recall and storage space usage. The Recall is average recall (\%) at top 1 (@1), 5 (@5) and 1\% (@1\%) test on KITTI dataset, the query size is average query point cloud size and the map size is total size of maps.}
\label{tab:recall}

\begin{tabular}{@{}clccccc@{}}
\toprule
                                   Category  & Method           & Recall@1($\uparrow$) & Recall@5($\uparrow$) & Recall@1\%($\uparrow$) & query size($\downarrow$)                & map size($\downarrow$)                  \\ \midrule
\multirow{4}{*}{Lidar to Uncompressed map}   & PointNetVLAD~\cite{uy2018pointnetvlad}     & 88.69    & 97.17    & 89.53      & \multirow{4}{*}{14.22 MB} & \multirow{4}{*}{\textcolor{blue}{18161.15 MB}}  \\
                                             & LPD-Net~\cite{liu2019lpd}          & \textcolor{blue}{89.54}    & \textbf{99.78}    & \textbf{93.24}      &                           &                           \\
                                             & MinkLoc3D~\cite{Minkloc3d}        & 21.83    & 60.85    & 26.54      &                           &                           \\
                                            \midrule
\multicolumn{1}{c}{Lidar to Compressed map} & Retriever~\cite{wiesmann2022retriever}        & 82.06    & 95.88    & 86.27      & 14.22 MB                  & \textbf{59.81 MB}                  \\ \midrule
\multirow{3}{*}{Images to Compressed map}    & $\text{VOLoc}_{DSO}$(ours)  & \textbf{91.01}    & \textcolor{blue}{99.17}    & \textcolor{blue}{91.82}      & 2.89 MB                   & \multirow{3}{*}{\textbf{59.81 MB}} \\
                                             & $\text{VOLoc}_{VINS-Mono}$(ours) & 85.7     & 95.35    & 85.7       & \textcolor{blue}{0.26 MB}                   &                           \\
                                             & $\text{VOLoc}_{ORB-SLAM3}$(ours)  & 72.62    & 88.17    & 75.96      & \textbf{0.03 MB}                  &                           \\ \bottomrule
\end{tabular}
\vspace{-2 em}
\end{table*}

\subsection{Dataset}

For the lack of an available dataset to evaluate Image-to-Compressed Lidar map VPR, we spent months constructing a testing dataset based on KITTI~\cite{Geiger2013VisionMR}. The KITTI dataset is a well-known large-scale autonomous driving dataset. It has multiple sensor data like cameras, Lidar, IMU and GPS. We use eleven sequences (00-10) in the KITTI Visual Odometry dataset to 
construct the compressed maps using the Lidar data and conduct VO-based query using the camera data. 

Following Wiesmann~\cite{wiesmann2021deep}, we first build whole Lidar maps by groud truth poses for each sequence and divide them into $40 \times 40 \times 15~m^3$ sub-maps at $20m$ intervals, and downsample by $0.1~m$ voxel grids. The global coordinates of each sub-map are the ground truth coordinates at its own centroid. 
We consider positive samples to be less than $20m$ apart from the anchor and the negative samples to be at least $50m$ away.
Sequences 07 and 10 are used for testing, and the rest are used for training. Due to challenging environments, DSO fails to run sequence 01 and VINS-mono fails to run sequence 00 and 03. We use the rest for training.

\subsection{Evaluation metrics}
For place recognition methods, $recall@K$ is the most widely used evaluation metric~\cite{uy2018pointnetvlad,liu2019lpd,zhang2019pcan, Minkloc3d}. If the retrieved top $K$ descriptors have at least one correct sample, then this retrieval is considered correct. We set $K$ to the $1\%$ of the database to make the metrics invariant to the size of the database~\cite{cattaneo2020global}. Following the settings of PointNetVLAD~\cite{uy2018pointnetvlad}, we consider a query is successfully localized if the retrieved submap is within $25m$.

\subsection{Localization Results}

The first experiment shows the localization performance and the storage space usage. Table~\ref{tab:recall} reports Recall@1, Recall@5, and Recall@1\%, the average query sub-map size, and total map size. 

We compare our method with two categories of methods: the Lidar to Uncompressed map~(LtoU) methods~\cite{uy2018pointnetvlad, liu2019lpd, Minkloc3d} and Lidar to Compressed map~(LtoC) method~\cite{wiesmann2022retriever}. The LtoU methods use a Lidar point cloud to query Uncompressed maps. The LtoC method retrieves a point cloud in Compressed maps. Our method queries Images in Compressed maps. For all compared methods, we use the official codes. We just modify the data loading codes to fit our KITTI dataset and train them with the default configuration. Due to the lack of publicly available codes, we cannot compare with Image to Uncompressed map methods~\cite{I2P-Rec, cattaneo2020global}.

\begin{figure}[ht]
  \centering
  \includegraphics[width=0.9\linewidth]{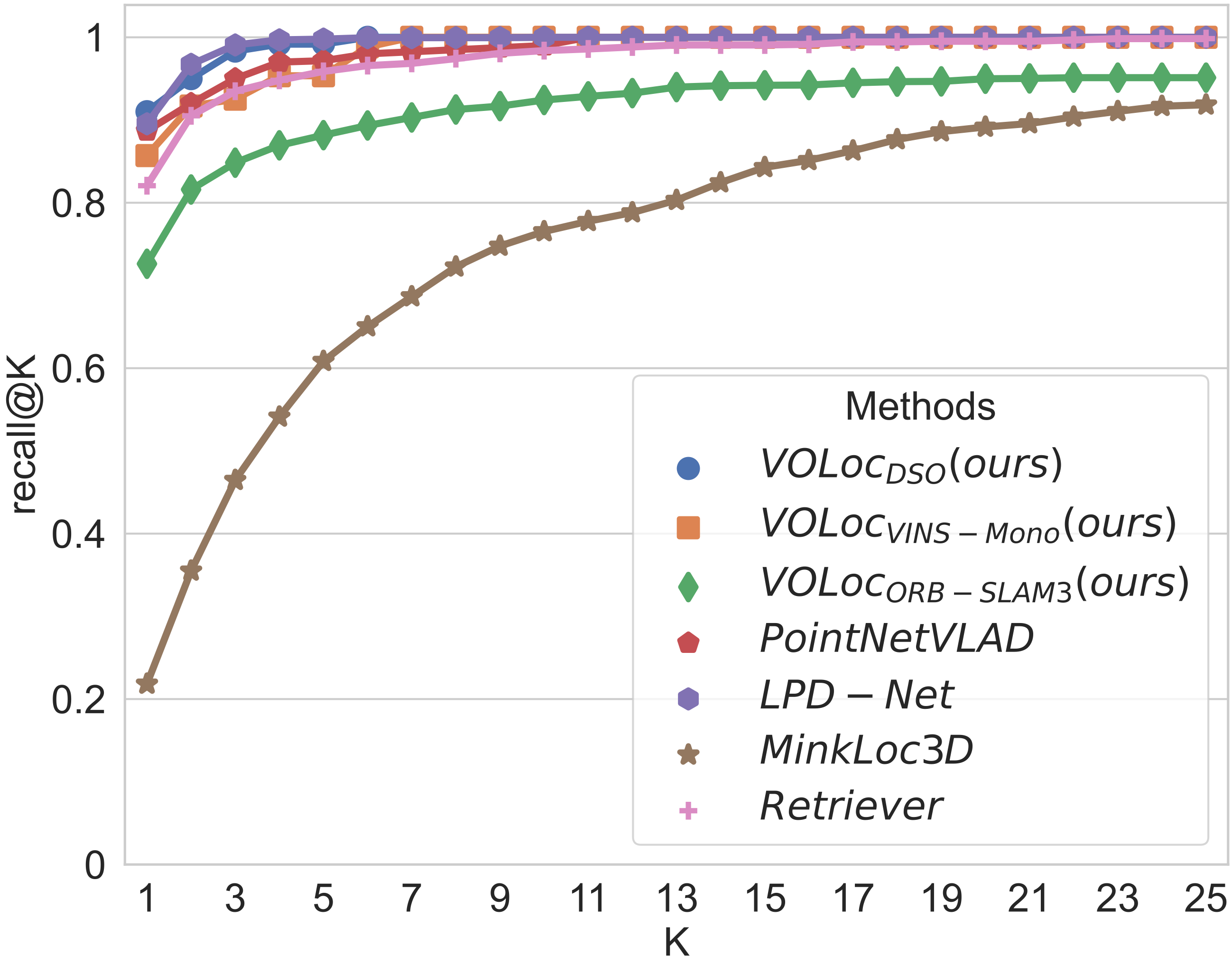}
  \caption{Average recall@K on the KITTI dataset.}
  \label{fig:compare}
\vspace{-1.5 em}
\end{figure}

Our best model ($\text{VOLoc}_{DSO}$) outperforms most of the baseline methods and slightly inferiors LPD-Net, which performs the best on the KITTI dataset. Other two models ($\text{VOLoc}_{VINS-Mono}$ and $\text{VOLoc}_{ORB-SLAM3}$) also achieve comparable performance with other baseline methods. However, it is not a fair comparison as all methods except Retriever~\cite{wiesmann2022retriever} query a Lidar-based point cloud in the uncompressed point clouds. Table~\ref{tab:recall} reveals that the query size and map size of our method is much smaller than Lidar to Uncompressed map methods. Compared to Retriever, our $\text{VOLoc}_{DSO}$ and $\text{VOLoc}_{VINS-Mono}$ methods outperform in localization performance, with a much smaller query size. Consequently, the advantage of our methods is that we directly localize images in the compressed map with little extra space occupancy (the reconstruction visual point cloud). Our method is suitable for mobile devices with limited storage space and transmission bandwidth. 
Figure~\ref{fig:compare} shows the average recall @K on the KITTI dataset. 

\subsection{Ablation Studies}
In this section, we investigate the effect of different system components. Note that the "base" model means it was trained from scratch without transfer learning and Visual Point Clouds Refine and only uses $L_{VtoL}$ in the loss function.

\paragraph{Transfer learning}
As shown in Table~\ref{tab:ablation}, the base model performs unsatisfactorily regardless of the VO. With the proposed transfer learning strategy, the Recall@1 is improved by 11.47\%, 7.33\% and 11.47\%, respectively. The transfer learning makes the model learn more geometric features, enhancing the performance of all the VO methods.

\paragraph{Combined Loss}
The combined Loss enhances localization performance, as shown in table~\ref{tab:ablation}. This implies that making the descriptors of QPCs more discriminative helps the network find a better correlation between visual and Lidar point clouds. The Recall@1 of DSO-based method has been enhanced to 91.34\% and surpasses LPD-Net (89.54\%). The other two methods also are improved by (7.83\% and 3.97\%).

\paragraph{Visual Point Cloud Refine} 
Point clouds from ORB-SLAM3 and VINS-Mono are more sparse than those of DSO.Table~\ref{tab:ablation} shows that the optimization significantly narrows performance gaps. The optimization has a huge boost to the ORB-SLAM3-based method. Its Recall@1 reaches 72.62\%. It also enhances the VINS-Mono-based method by 6.82\%, but has a slight effect on the DSO-based method due to its inherent density.
\begin{table}
 \caption{Ablation Studies. VO means which VO is used to construct the query. TL refers to Transfer learning.}
\label{tab:ablation}
\begin{tabular}{@{}cccccc@{}}
\toprule
VO                         & \multicolumn{1}{c}{base} & \multicolumn{1}{c}{TL} & \multicolumn{1}{c}{combined loss} & \multicolumn{1}{c}{QPC Refine} & Recall@1 \\ \midrule
\multirow{4}{*}{DSO}       & $\checkmark$             &                               &                              &                             & 74.38     \\
                           & $\checkmark$             & $\checkmark$                  &                              &                             & 85.85(+11.47)     \\
                           & $\checkmark$             & $\checkmark$                  & $\checkmark$                 &                             & \textcolor{blue}{91.34(+5.49)}     \\
                           & $\checkmark$             & $\checkmark$                  & $\checkmark$                 & $\checkmark$                & \textbf{91.84(+0.5)}     \\ \midrule
\multirow{4}{*}{VINS-Mono} & $\checkmark$             &                               &                              &                             & 63.72     \\
                           & $\checkmark$             & $\checkmark$                  &                              &                             & 71.05(+7.33)     \\
                           & $\checkmark$             & $\checkmark$                  & $\checkmark$                 &                             & \textcolor{blue}{78.88(+7.83)}     \\
                           & $\checkmark$             & $\checkmark$                  & $\checkmark$                 & $\checkmark$                & \textbf{85.70(+6.82)}     \\ \midrule
\multirow{4}{*}{ORB-SLAM3}  & $\checkmark$             &                               &                              &                             & 42.29     \\
                           & $\checkmark$             & $\checkmark$                  &                              &                             & 53.76(+11.47)     \\
                           & $\checkmark$             & $\checkmark$                  & $\checkmark$                 &                             & \textcolor{blue}{57.73(+3.97)}     \\
                           & $\checkmark$             & $\checkmark$                  & $\checkmark$                 & $\checkmark$                & \textbf{72.62(+14.89)}     \\ \bottomrule
\end{tabular}
\end{table}

\subsection{Qualitative results and Visualization}
\begin{figure}[ht]
\vspace{-1.2 em}
  \centering
  \includegraphics[width=\linewidth]{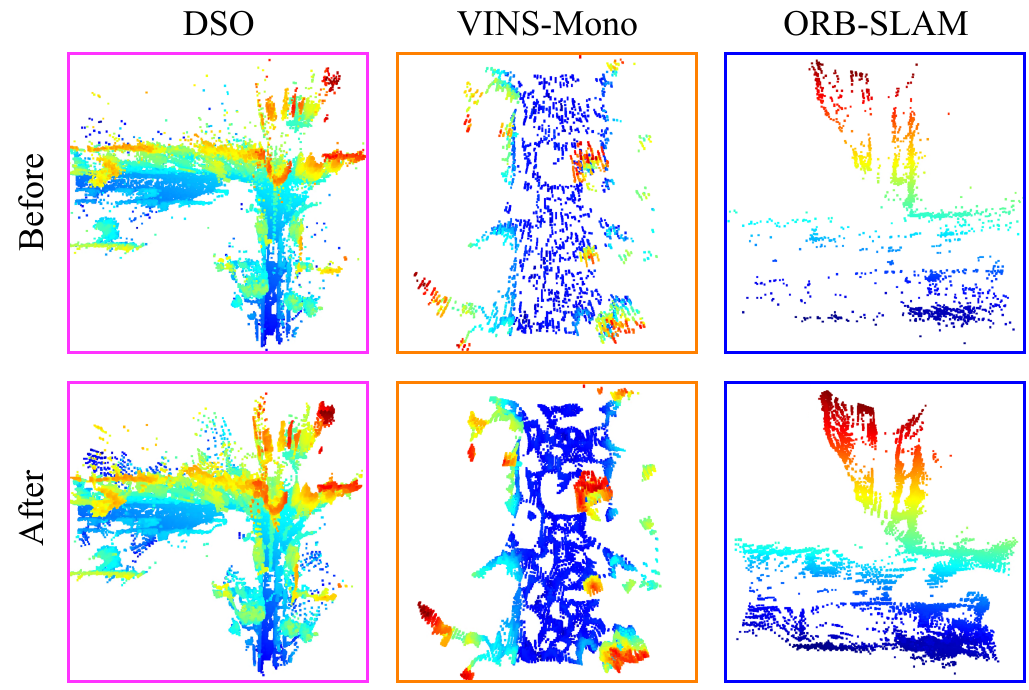}
  \caption{Point cloud optimization effects.}
  \label{fig:densification}
\vspace{-1 em}
\end{figure}

In this section, we show qualitative results of Visual Point Cloud Refine and retrieval results to provide a better understanding of our approach and the challenges of the task.
\paragraph{Visual Point Cloud Refine}
Figure~\ref{fig:densification} shows the effects of the Visual Point Cloud Refine~(section~\ref{sec:GRM}). It has a slight impact on the point clouds of DSO, but has obvious effects on the point clouds of VINS-Mono and ORB-SLAM3. These results are also reflected in the ablation experiments.

\begin{figure}[ht]
  \centering
  \includegraphics[width=0.95\linewidth]{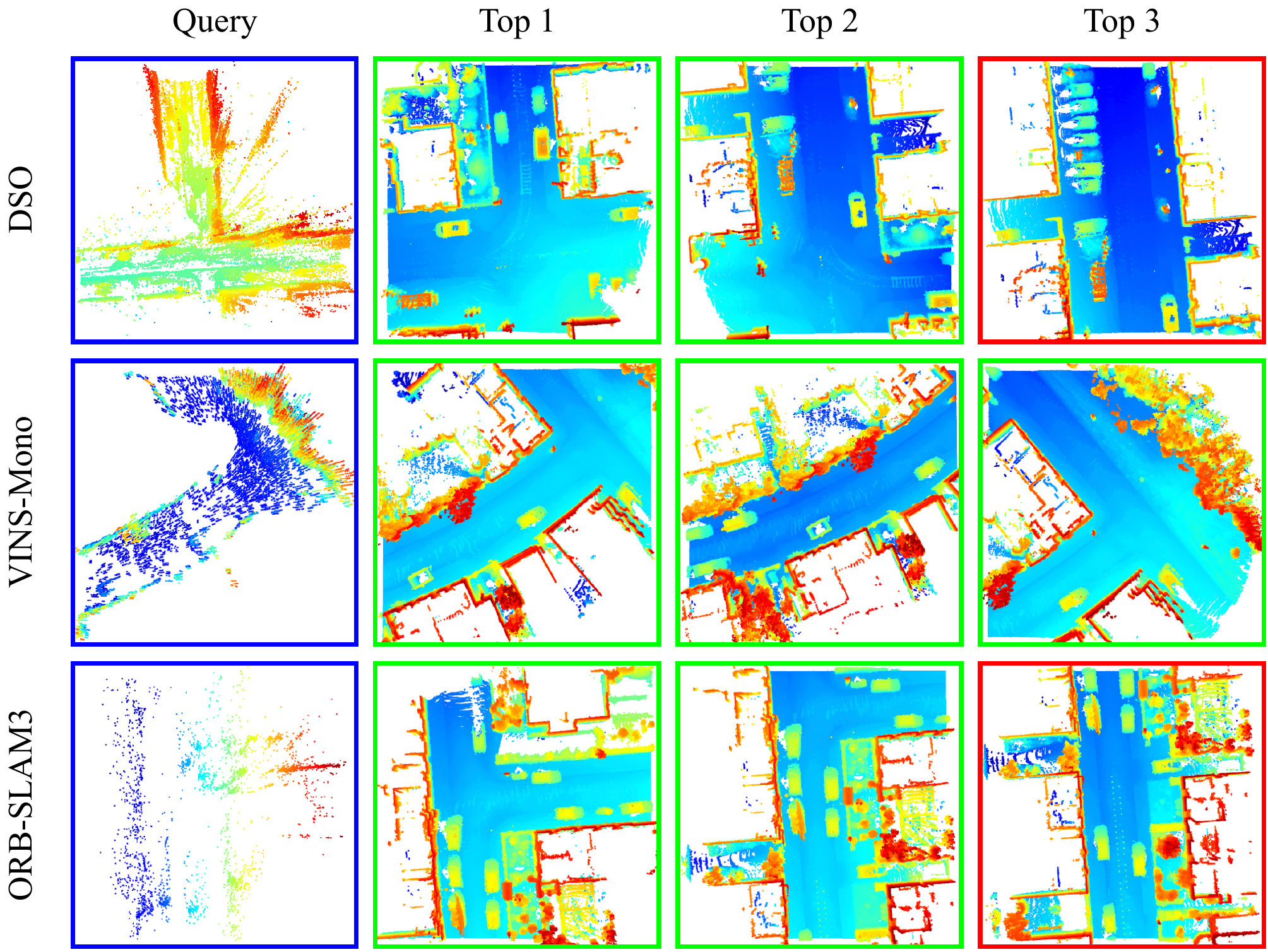}
  \caption{Top 3 retrieval results when different VO are used. The blue box refers to query, the green box means correct match, and the red one refers to a wrong match.}
  \label{fig:retrieval_results}
\end{figure}

\paragraph{Retrieval results}
Figure~\ref{fig:retrieval_results} demonstrates the top 3 retrieval results of three different VO methods. The displayed queries are not refined and the retrieved sub-maps are the Lidar sub-maps.
 The gap between Visual Point Clouds and Lidar sub-maps is noticeable, but our methods can work in most cases. However, the sub-maps from different locations may be similar, leading to false matches.

\vspace{-0.5 em}
\subsection{Time consumption}
\begin{figure}[ht]
\vspace{-1 em}
  \centering
  \includegraphics[width=0.9\linewidth]{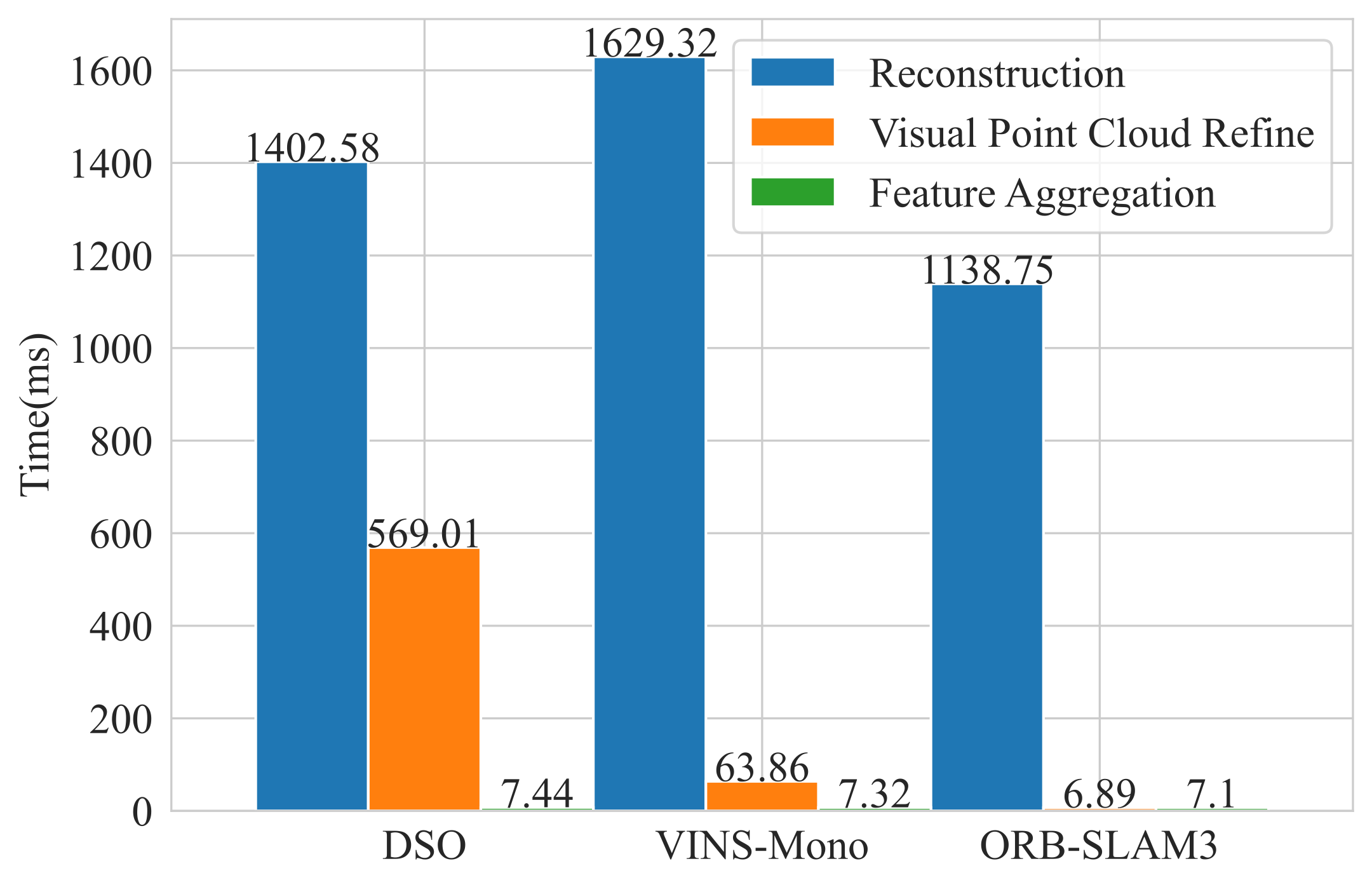}
  \caption{The average time consumption to process a visual sub-map of each part in VOLoc.}
  \label{fig:Time consumption}
\vspace{-1.5 em}
\end{figure}

We test the time cost of each part of our method, as shown in Figure~\ref{fig:Time consumption}. The time of our Visual Point cloud Refine and Feature Aggregation modules is much less than the time the VO takes to rebuild the sub-maps, which means our methods can carry out place recognition in real-time. The densification for sub-maps generated by DSO is slightly time-consuming, but it is highly efficient for sub-maps rebuilt by VINS-Mono and ORB-SLAM3 and obviously boosts the localization accuracy.

\section{Conclusion}
This paper presents VOLoc, which utilizes geometrical similarity to localize images in compressed Lidar maps. The proposed GRM module recovers geometric structure from images and refines it for better geometric quality. GPC is exploited to compress the Lidar maps while keeping the geometric consistency. A transfer learning scheme is proposed to train the attention-based aggregation network, which is crucial for the network to focus on the important points.
We evaluate our methods on the KITTI dataset and provide comprehensive experiments to validate the methods. The results show that the proposed methods are memory efficient and perform comparable to Lidar-to-Lidar place recognition methods. Despite the promising findings of this study, we acknowledge that our method can only handle sequence images, thus limiting the applications. Further research is needed to investigate how to utilize geometrical similarity with a single image. We believe that VOLoc provides a new way for the Image-to-Lidar place recognition task.

\bibliographystyle{IEEEtran}
\bibliography{References}

\begin{thebibliography}{10}
\providecommand{\url}[1]{#1}
\csname url@samestyle\endcsname
\providecommand{\newblock}{\relax}
\providecommand{\bibinfo}[2]{#2}
\providecommand{\BIBentrySTDinterwordspacing}{\spaceskip=0pt\relax}
\providecommand{\BIBentryALTinterwordstretchfactor}{4}
\providecommand{\BIBentryALTinterwordspacing}{\spaceskip=\fontdimen2\font plus
\BIBentryALTinterwordstretchfactor\fontdimen3\font minus
  \fontdimen4\font\relax}
\providecommand{\BIBforeignlanguage}[2]{{%
\expandafter\ifx\csname l@#1\endcsname\relax
\typeout{** WARNING: IEEEtran.bst: No hyphenation pattern has been}%
\typeout{** loaded for the language `#1'. Using the pattern for}%
\typeout{** the default language instead.}%
\else
\language=\csname l@#1\endcsname
\fi
#2}}
\providecommand{\BIBdecl}{\relax}
\BIBdecl

\bibitem{HLoc}
P.-E. Sarlin, C.~Cadena, R.~Siegwart, and M.~Dymczyk, ``From coarse to fine:
  Robust hierarchical localization at large scale,'' in \emph{Proceedings of
  the IEEE/CVF Conference on Computer Vision and Pattern Recognition}, 2019,
  pp. 12\,716--12\,725.

\bibitem{torii201524}
A.~Torii, R.~Arandjelovic, J.~Sivic, M.~Okutomi, and T.~Pajdla, ``24/7 place
  recognition by view synthesis,'' in \emph{Proceedings of the IEEE conference
  on computer vision and pattern recognition}, 2015, pp. 1808--1817.

\bibitem{DBLP:conf/cvpr/BrachmannR18}
\BIBentryALTinterwordspacing
E.~Brachmann and C.~Rother, ``Learning less is more - 6d camera localization
  via 3d surface regression,'' in \emph{2018 {IEEE} Conference on Computer
  Vision and Pattern Recognition, {CVPR} 2018, Salt Lake City, UT, USA, June
  18-22, 2018}.\hskip 1em plus 0.5em minus 0.4em\relax Computer Vision
  Foundation / {IEEE} Computer Society, 2018, pp. 4654--4662. [Online].
  Available:
  \url{http://openaccess.thecvf.com/content\_cvpr\_2018/html/Brachmann\_Learning\_Less\_Is\_CVPR\_2018\_paper.html}
\BIBentrySTDinterwordspacing

\bibitem{arandjelovic2016netvlad}
R.~Arandjelovic, P.~Gronat, A.~Torii, T.~Pajdla, and J.~Sivic, ``Netvlad: Cnn
  architecture for weakly supervised place recognition,'' in \emph{Proceedings
  of the IEEE conference on computer vision and pattern recognition}, 2016, pp.
  5297--5307.

\bibitem{DBLP:journals/ral/XieDWLWTZ22}
\BIBentryALTinterwordspacing
T.~Xie, K.~Dai, K.~Wang, R.~Li, J.~Wang, X.~Tang, and L.~Zhao, ``A deep feature
  aggregation network for accurate indoor camera localization,'' \emph{{IEEE}
  Robotics Autom. Lett.}, vol.~7, no.~2, pp. 3687--3694, 2022. [Online].
  Available: \url{https://doi.org/10.1109/LRA.2022.3146946}
\BIBentrySTDinterwordspacing

\bibitem{sattler2016efficient}
T.~Sattler, B.~Leibe, and L.~Kobbelt, ``Efficient \& effective prioritized
  matching for large-scale image-based localization,'' \emph{IEEE transactions
  on pattern analysis and machine intelligence}, vol.~39, no.~9, pp.
  1744--1756, 2016.

\bibitem{toft2020long}
C.~Toft, W.~Maddern, A.~Torii, L.~Hammarstrand, E.~Stenborg, D.~Safari,
  M.~Okutomi, M.~Pollefeys, J.~Sivic, T.~Pajdla \emph{et~al.}, ``Long-term
  visual localization revisited,'' \emph{IEEE Transactions on Pattern Analysis
  and Machine Intelligence}, vol.~44, no.~4, pp. 2074--2088, 2020.

\bibitem{Nuscenes}
H.~Caesar, V.~Bankiti, A.~H. Lang, S.~Vora, V.~E. Liong, Q.~Xu, A.~Krishnan,
  Y.~Pan, G.~Baldan, and O.~Beijbom, ``nuscenes: A multimodal dataset for
  autonomous driving,'' in \emph{Proceedings of the IEEE/CVF conference on
  computer vision and pattern recognition}, 2020, pp. 11\,621--11\,631.

\bibitem{Waymo}
\BIBentryALTinterwordspacing
P.~Sun, H.~Kretzschmar, X.~Dotiwalla, A.~Chouard, V.~Patnaik, P.~Tsui, J.~Guo,
  Y.~Zhou, Y.~Chai, B.~Caine, V.~Vasudevan, W.~Han, J.~Ngiam, H.~Zhao,
  A.~Timofeev, S.~M. Ettinger, M.~Krivokon, A.~Gao, A.~Joshi, Y.~Zhang,
  J.~Shlens, Z.~Chen, and D.~Anguelov, ``Scalability in perception for
  autonomous driving: Waymo open dataset,'' \emph{2020 IEEE/CVF Conference on
  Computer Vision and Pattern Recognition (CVPR)}, pp. 2443--2451, 2019.
  [Online]. Available: \url{https://api.semanticscholar.org/CorpusID:209140225}
\BIBentrySTDinterwordspacing

\bibitem{cattaneo2020global}
D.~Cattaneo, M.~Vaghi, S.~Fontana, A.~L. Ballardini, and D.~G. Sorrenti,
  ``Global visual localization in lidar-maps through shared 2d-3d embedding
  space,'' in \emph{2020 IEEE International Conference on Robotics and
  Automation (ICRA)}.\hskip 1em plus 0.5em minus 0.4em\relax IEEE, 2020, pp.
  4365--4371.

\bibitem{DBLP:conf/cvpr/YangSLL0CT22}
\BIBentryALTinterwordspacing
L.~Yang, R.~Shrestha, W.~Li, S.~Liu, G.~Zhang, Z.~Cui, and P.~Tan,
  ``Scenesqueezer: Learning to compress scene for camera relocalization,'' in
  \emph{{IEEE/CVF} Conference on Computer Vision and Pattern Recognition,
  {CVPR} 2022, New Orleans, LA, USA, June 18-24, 2022}.\hskip 1em plus 0.5em
  minus 0.4em\relax {IEEE}, 2022, pp. 8249--8258. [Online]. Available:
  \url{https://doi.org/10.1109/CVPR52688.2022.00808}
\BIBentrySTDinterwordspacing

\bibitem{I2P-Rec}
\BIBentryALTinterwordspacing
Y.~Li, S.~Zheng, Z.~Yu, B.~Yu, S.~Cao, L.~Luo, and H.~Shen, ``I2p-rec:
  Recognizing images on large-scale point cloud maps through bird's eye view
  projections,'' \emph{ArXiv}, vol. abs/2303.01043, 2023. [Online]. Available:
  \url{https://api.semanticscholar.org/CorpusID:257280108}
\BIBentrySTDinterwordspacing

\bibitem{DBLP:conf/mir/SunL0F020}
\BIBentryALTinterwordspacing
Q.~Sun, H.~Liu, J.~He, Z.~Fan, and X.~Du, ``{DAGC:} employing dual attention
  and graph convolution for point cloud based place recognition,'' in
  \emph{Proceedings of the 2020 on International Conference on Multimedia
  Retrieval, {ICMR} 2020, Dublin, Ireland, June 8-11, 2020}, C.~Gurrin,
  B.~{\TH}. J{\'{o}}nsson, N.~Kando, K.~Sch{\"{o}}ffmann, Y.~P. Chen, and N.~E.
  O'Connor, Eds.\hskip 1em plus 0.5em minus 0.4em\relax {ACM}, 2020, pp.
  224--232. [Online]. Available: \url{https://doi.org/10.1145/3372278.3390693}
\BIBentrySTDinterwordspacing

\bibitem{uy2018pointnetvlad}
M.~A. Uy and G.~H. Lee, ``Pointnetvlad: Deep point cloud based retrieval for
  large-scale place recognition,'' in \emph{Proceedings of the IEEE conference
  on computer vision and pattern recognition}, 2018, pp. 4470--4479.

\bibitem{DBLP:conf/cvpr/WeiBWMU19}
\BIBentryALTinterwordspacing
X.~Wei, I.~A. Barsan, S.~Wang, J.~Martinez, and R.~Urtasun, ``Learning to
  localize through compressed binary maps,'' in \emph{{IEEE} Conference on
  Computer Vision and Pattern Recognition, {CVPR} 2019, Long Beach, CA, USA,
  June 16-20, 2019}.\hskip 1em plus 0.5em minus 0.4em\relax Computer Vision
  Foundation / {IEEE}, 2019, pp. 10\,316--10\,324. [Online]. Available:
  \url{http://openaccess.thecvf.com/content\_CVPR\_2019/html/Wei\_Learning\_to\_Localize\_Through\_Compressed\_Binary\_Maps\_CVPR\_2019\_paper.html}
\BIBentrySTDinterwordspacing

\bibitem{wiesmann2022retriever}
L.~Wiesmann, R.~Marcuzzi, C.~Stachniss, and J.~Behley, ``Retriever: Point cloud
  retrieval in compressed 3d maps,'' in \emph{2022 International Conference on
  Robotics and Automation (ICRA)}.\hskip 1em plus 0.5em minus 0.4em\relax IEEE,
  2022, pp. 10\,925--10\,932.

\bibitem{engel2017direct}
J.~Engel, V.~Koltun, and D.~Cremers, ``Direct sparse odometry,'' \emph{IEEE
  transactions on pattern analysis and machine intelligence}, vol.~40, no.~3,
  pp. 611--625, 2017.

\bibitem{campos2021orb}
C.~Campos, R.~Elvira, J.~J.~G. Rodr{\'\i}guez, J.~M. Montiel, and J.~D.
  Tard{\'o}s, ``Orb-slam3: An accurate open-source library for visual,
  visual--inertial, and multimap slam,'' \emph{IEEE Transactions on Robotics},
  vol.~37, no.~6, pp. 1874--1890, 2021.

\bibitem{qin2018vins}
T.~Qin, P.~Li, and S.~Shen, ``Vins-mono: A robust and versatile monocular
  visual-inertial state estimator,'' \emph{IEEE Transactions on Robotics},
  vol.~34, no.~4, pp. 1004--1020, 2018.

\bibitem{jaegle2021perceiver}
A.~Jaegle, F.~Gimeno, A.~Brock, O.~Vinyals, A.~Zisserman, and J.~Carreira,
  ``Perceiver: General perception with iterative attention,'' in
  \emph{International conference on machine learning}.\hskip 1em plus 0.5em
  minus 0.4em\relax PMLR, 2021, pp. 4651--4664.

\bibitem{Geiger2013VisionMR}
A.~Geiger, P.~Lenz, C.~Stiller, and R.~Urtasun, ``Vision meets robotics: The
  kitti dataset,'' \emph{The International Journal of Robotics Research},
  vol.~32, pp. 1231 -- 1237, 2013.

\bibitem{lowe2004distinctive}
D.~G. Lowe, ``Distinctive image features from scale-invariant keypoints,''
  \emph{International journal of computer vision}, vol.~60, pp. 91--110, 2004.

\bibitem{detone2018superpoint}
D.~DeTone, T.~Malisiewicz, and A.~Rabinovich, ``Superpoint: Self-supervised
  interest point detection and description,'' in \emph{Proceedings of the IEEE
  conference on computer vision and pattern recognition workshops}, 2018, pp.
  224--236.

\bibitem{sivic2003video}
J.~Sivic and A.~Zisserman, ``Video google: A text retrieval approach to object
  matching in videos,'' in \emph{Computer Vision, IEEE International Conference
  on}, vol.~3.\hskip 1em plus 0.5em minus 0.4em\relax IEEE Computer Society,
  2003, pp. 1470--1470.

\bibitem{jegou2010aggregating}
H.~J{\'e}gou, M.~Douze, C.~Schmid, and P.~P{\'e}rez, ``Aggregating local
  descriptors into a compact image representation,'' in \emph{2010 IEEE
  computer society conference on computer vision and pattern
  recognition}.\hskip 1em plus 0.5em minus 0.4em\relax IEEE, 2010, pp.
  3304--3311.

\bibitem{berton2022rethinking}
G.~Berton, C.~Masone, and B.~Caputo, ``Rethinking visual geo-localization for
  large-scale applications,'' in \emph{Proceedings of the IEEE/CVF Conference
  on Computer Vision and Pattern Recognition}, 2022, pp. 4878--4888.

\bibitem{APANet}
Y.~Zhu, J.~Wang, L.~Xie, and L.~Zheng, ``Attention-based pyramid aggregation
  network for visual place recognition,'' in \emph{Proceedings of the 26th ACM
  international conference on Multimedia}, 2018, pp. 99--107.

\bibitem{DMPCANet}
Y.~Wang, H.~Chen, J.~Wang, and Y.~Zhu, ``Dmpcanet: A low dimensional
  aggregation network for visual place recognition,'' in \emph{Proceedings of
  the 2022 International Conference on Multimedia Retrieval}, 2022, pp. 24--28.

\bibitem{vgg}
K.~Simonyan and A.~Zisserman, ``Very deep convolutional networks for
  large-scale image recognition,'' \emph{arXiv preprint arXiv:1409.1556}, 2014.

\bibitem{resnet}
K.~He, X.~Zhang, S.~Ren, and J.~Sun, ``Deep residual learning for image
  recognition,'' in \emph{Proceedings of the IEEE conference on computer vision
  and pattern recognition}, 2016, pp. 770--778.

\bibitem{hausler2021patch}
S.~Hausler, S.~Garg, M.~Xu, M.~Milford, and T.~Fischer, ``Patch-netvlad:
  Multi-scale fusion of locally-global descriptors for place recognition,'' in
  \emph{Proceedings of the IEEE/CVF Conference on Computer Vision and Pattern
  Recognition}, 2021, pp. 14\,141--14\,152.

\bibitem{jin2017learned}
H.~Jin~Kim, E.~Dunn, and J.-M. Frahm, ``Learned contextual feature reweighting
  for image geo-localization,'' in \emph{Proceedings of the IEEE Conference on
  Computer Vision and Pattern Recognition}, 2017, pp. 2136--2145.

\bibitem{peng2021semantic}
G.~Peng, Y.~Yue, J.~Zhang, Z.~Wu, X.~Tang, and D.~Wang, ``Semantic reinforced
  attention learning for visual place recognition,'' in \emph{2021 IEEE
  International Conference on Robotics and Automation (ICRA)}.\hskip 1em plus
  0.5em minus 0.4em\relax IEEE, 2021, pp. 13\,415--13\,422.

\bibitem{ScanContext}
\BIBentryALTinterwordspacing
G.~Kim and A.~Kim, ``Scan context: Egocentric spatial descriptor for place
  recognition within 3d point cloud map,'' \emph{2018 IEEE/RSJ International
  Conference on Intelligent Robots and Systems (IROS)}, pp. 4802--4809, 2018.
  [Online]. Available: \url{https://api.semanticscholar.org/CorpusID:57755993}
\BIBentrySTDinterwordspacing

\bibitem{ScanContext++}
\BIBentryALTinterwordspacing
G.~Kim, S.~Choi, and A.~Kim, ``Scan context++: Structural place recognition
  robust to rotation and lateral variations in urban environments,'' \emph{IEEE
  Transactions on Robotics}, vol.~38, pp. 1856--1874, 2021. [Online].
  Available: \url{https://api.semanticscholar.org/CorpusID:238198272}
\BIBentrySTDinterwordspacing

\bibitem{DiSCO}
\BIBentryALTinterwordspacing
X.~Xu, H.~Yin, Z.~Chen, Y.~Li, Y.~Wang, and R.~Xiong, ``Disco: Differentiable
  scan context with orientation,'' \emph{IEEE Robotics and Automation Letters},
  vol.~6, pp. 2791--2798, 2020. [Online]. Available:
  \url{https://api.semanticscholar.org/CorpusID:224814531}
\BIBentrySTDinterwordspacing

\bibitem{BEVPlace}
\BIBentryALTinterwordspacing
L.~Luo, S.~Zheng, Y.~Li, Y.~H. Fan, B.~Yu, S.~Cao, and H.~Shen, ``Bevplace:
  Learning lidar-based place recognition using bird's eye view images,''
  \emph{ArXiv}, vol. abs/2302.14325, 2023. [Online]. Available:
  \url{https://api.semanticscholar.org/CorpusID:257232932}
\BIBentrySTDinterwordspacing

\bibitem{qi2017pointnet}
C.~R. Qi, H.~Su, K.~Mo, and L.~J. Guibas, ``Pointnet: Deep learning on point
  sets for 3d classification and segmentation,'' in \emph{Proceedings of the
  IEEE conference on computer vision and pattern recognition}, 2017, pp.
  652--660.

\bibitem{liu2019lpd}
Z.~Liu, S.~Zhou, C.~Suo, P.~Yin, W.~Chen, H.~Wang, H.~Li, and Y.-H. Liu,
  ``Lpd-net: 3d point cloud learning for large-scale place recognition and
  environment analysis,'' in \emph{Proceedings of the IEEE/CVF International
  Conference on Computer Vision}, 2019, pp. 2831--2840.

\bibitem{zhang2019pcan}
W.~Zhang and C.~Xiao, ``Pcan: 3d attention map learning using contextual
  information for point cloud based retrieval,'' in \emph{Proceedings of the
  IEEE/CVF Conference on Computer Vision and Pattern Recognition}, 2019, pp.
  12\,436--12\,445.

\bibitem{sun2020dagc}
Q.~Sun, H.~Liu, J.~He, Z.~Fan, and X.~Du, ``Dagc: Employing dual attention and
  graph convolution for point cloud based place recognition,'' in
  \emph{Proceedings of the 2020 International Conference on Multimedia
  Retrieval}, 2020, pp. 224--232.

\bibitem{Minkloc3d}
J.~Komorowski, ``Minkloc3d: Point cloud based large-scale place recognition,''
  in \emph{2021 IEEE Winter Conference on Applications of Computer Vision
  (WACV)}, 2021, pp. 1789--1798.

\bibitem{MinkLoc3Dv2}
\BIBentryALTinterwordspacing
------, ``Improving point cloud based place recognition with ranking-based loss
  and large batch training,'' in \emph{26th International Conference on Pattern
  Recognition, {ICPR} 2022, Montreal, QC, Canada, August 21-25, 2022}.\hskip
  1em plus 0.5em minus 0.4em\relax {IEEE}, 2022, pp. 3699--3705. [Online].
  Available: \url{https://doi.org/10.1109/ICPR56361.2022.9956458}
\BIBentrySTDinterwordspacing

\bibitem{OverlapTransformer}
\BIBentryALTinterwordspacing
J.~Ma, J.~Zhang, J.~Xu, R.~Ai, W.~Gu, and X.~Chen, ``Overlaptransformer: An
  efficient and yaw-angle-invariant transformer network for lidar-based place
  recognition,'' \emph{IEEE Robotics and Automation Letters}, vol.~7, pp.
  6958--6965, 2022. [Online]. Available:
  \url{https://api.semanticscholar.org/CorpusID:249223069}
\BIBentrySTDinterwordspacing

\bibitem{wolcott2014visual}
R.~W. Wolcott and R.~M. Eustice, ``Visual localization within lidar maps for
  automated urban driving,'' in \emph{2014 IEEE/RSJ International Conference on
  Intelligent Robots and Systems}.\hskip 1em plus 0.5em minus 0.4em\relax IEEE,
  2014, pp. 176--183.

\bibitem{cattaneo2019cmrnet}
D.~Cattaneo, M.~Vaghi, A.~L. Ballardini, S.~Fontana, D.~G. Sorrenti, and
  W.~Burgard, ``Cmrnet: Camera to lidar-map registration,'' in \emph{2019 IEEE
  intelligent transportation systems conference (ITSC)}.\hskip 1em plus 0.5em
  minus 0.4em\relax IEEE, 2019, pp. 1283--1289.

\bibitem{caselitz2016monocular}
T.~Caselitz, B.~Steder, M.~Ruhnke, and W.~Burgard, ``Monocular camera
  localization in 3d lidar maps,'' in \emph{2016 IEEE/RSJ International
  Conference on Intelligent Robots and Systems (IROS)}.\hskip 1em plus 0.5em
  minus 0.4em\relax IEEE, 2016, pp. 1926--1931.

\bibitem{li20203d}
Q.~Li, J.~Zhu, J.~Liu, R.~Cao, H.~Fu, J.~M. Garibaldi, Q.~Li, B.~Liu, and
  G.~Qiu, ``3d map-guided single indoor image localization refinement,''
  \emph{ISPRS Journal of Photogrammetry and Remote Sensing}, vol. 161, pp.
  13--26, 2020.

\bibitem{feng20192d3d}
M.~Feng, S.~Hu, M.~H. Ang, and G.~H. Lee, ``2d3d-matchnet: Learning to match
  keypoints across 2d image and 3d point cloud,'' in \emph{2019 International
  Conference on Robotics and Automation (ICRA)}.\hskip 1em plus 0.5em minus
  0.4em\relax IEEE, 2019, pp. 4790--4796.

\bibitem{zhong2009intrinsic}
Y.~Zhong, ``Intrinsic shape signatures: A shape descriptor for 3d object
  recognition,'' in \emph{2009 IEEE 12th international conference on computer
  vision workshops, ICCV workshops}.\hskip 1em plus 0.5em minus 0.4em\relax
  IEEE, 2009, pp. 689--696.

\bibitem{pham2020lcd}
Q.-H. Pham, M.~A. Uy, B.-S. Hua, D.~T. Nguyen, G.~Roig, and S.-K. Yeung, ``Lcd:
  Learned cross-domain descriptors for 2d-3d matching,'' in \emph{Proceedings
  of the AAAI Conference on Artificial Intelligence}, vol.~34, no.~07, 2020,
  pp. 11\,856--11\,864.

\bibitem{wiesmann2021deep}
L.~Wiesmann, A.~Milioto, X.~Chen, C.~Stachniss, and J.~Behley, ``Deep
  compression for dense point cloud maps,'' \emph{IEEE Robotics and Automation
  Letters}, vol.~6, no.~2, pp. 2060--2067, 2021.

\bibitem{KPConv}
\BIBentryALTinterwordspacing
H.~Thomas, C.~Qi, J.-E. Deschaud, B.~Marcotegui, F.~Goulette, and L.~J. Guibas,
  ``Kpconv: Flexible and deformable convolution for point clouds,'' \emph{2019
  IEEE/CVF International Conference on Computer Vision (ICCV)}, pp. 6410--6419,
  2019. [Online]. Available:
  \url{https://api.semanticscholar.org/CorpusID:121328056}
\BIBentrySTDinterwordspacing

\bibitem{maddern20171}
W.~Maddern, G.~Pascoe, C.~Linegar, and P.~Newman, ``1 year, 1000 km: The oxford
  robotcar dataset,'' \emph{The International Journal of Robotics Research},
  vol.~36, no.~1, pp. 3--15, 2017.

\end{thebibliography}
\end{document}